\documentclass[10pt]{article}

\usepackage[preprint]{tmlr}

\usepackage{amsmath,amsfonts,bm}

\def\eqref#1{equation~\ref{#1}}

\def\1{\bm{1}}

\DeclareMathAlphabet{\mathsfit}{\encodingdefault}{\sfdefault}{m}{sl}
\SetMathAlphabet{\mathsfit}{bold}{\encodingdefault}{\sfdefault}{bx}{n}

\newcommand{\E}{\mathbb{E}}

\newcommand{\R}{\mathbb{R}}

\newcommand{\Var}{\mathrm{Var}}

\newcommand{\Cov}{\mathrm{Cov}}

\usepackage{hyperref}     
\usepackage{thmtools}     
\usepackage{amssymb}      
\usepackage{booktabs}     
\usepackage{array}        
\usepackage{colortbl}     
\usepackage{longtable}    
\usepackage{enumitem}     
\usepackage{pgfplots}     
\usepackage{graphicx}     
\usepackage[T1]{fontenc}  
\usepackage{microtype}    
\usepackage{titlesec}     
\usepackage{bookmark}     
\usepackage{tikz}         
\usepackage{xcolor}       

\usetikzlibrary{positioning, arrows.meta, shapes.geometric, shadows.blur, fit}

\definecolor{linkcolor}{RGB}{83,83,182}
\definecolor{blue}{RGB}{25,113,194}
\definecolor{red}{RGB}{224,49,49}
\definecolor{captionblue}{HTML}{4E79A7}
\definecolor{captionred}{HTML}{E15759}
\definecolor{boxblue}{HTML}{4E79A7}
\definecolor{boxorange}{HTML}{F28E2B}

\hypersetup{colorlinks=true, citecolor=linkcolor, linkcolor=linkcolor}

\newtheorem{theorem}{Theorem}

\declaretheorem[name=Assumption]{assumption}
\declaretheorem[name=Proof]{proof}
\declaretheorem[name=Corollary]{corollary}
\newenvironment{sketchproof}{\par\noindent\textit{Proof sketch.}\ }
{\hfill$\square$\par}

\newcommand{\hideTheoremLikeNumbers}{
  \renewcommand{\thetheorem}{}
  \renewcommand{\theassumption}{}
  \renewcommand{\thecorollary}{}
}
\newcommand{\restoreTheoremLikeNumbers}{
  \renewcommand{\thetheorem}{\arabic{theorem}}
  \renewcommand{\theassumption}{\arabic{assumption}}
  \renewcommand{\thecorollary}{\arabic{corollary}}
}

\newenvironment{theoremwithtag}[1]
{\begingroup\renewcommand{\thetheorem}{#1}\begin{theorem}}
    {\end{theorem}\endgroup}

\newenvironment{corollarywithtag}[1]
{\begingroup\renewcommand{\thecorollary}{#1}\begin{corollary}}
    {\end{corollary}\endgroup}

\newcommand{\eg}{e.g. \,}
\newcommand{\ie}{i.e.\,}

\titleclass{\subsubsubsection}{straight}[\subsubsection]
\newcounter{subsubsubsection}[subsubsection]
\renewcommand\thesubsubsubsection{\thesubsubsection.\arabic{subsubsubsection}}

\titleformat{\subsubsubsection}[runin]
{\normalfont\small\itshape}
{\thesubsubsubsection}
{0.5em}
{}[.]

\titlespacing*{\subsubsubsection}{0pt}{0.5ex}{0.8em}

\makeatletter
\providecommand{\toclevel@subsubsubsection}{4}

\makeatother

\usepackage{hyperref}
\usepackage{url}

\title{Post-Training Corrections \\ for Improved Time-Series Forecasting}

\author{\name Hamza Cherkaoui \email hamza.cherkaoui@telecom-sudparis.eu \\
      \addr SAMOVAR\\
    Télécom SudParis\\
    Institut Polytechnique de Paris\\
    91120 Palaiseau
      \AND
      \name Malik Tiomoko \email malik.tiomoko@huawei.com \\
      \addr Noah Ark Lab\\
    Boulogne Billancourt\\
    France
      \AND
      \name Giuseppe Paolo \email giuseppe.paolo@huawei.com \\
      \addr Noah Ark Lab\\
    Boulogne Billancourt\\
    France
      \AND
      \name Yili Zhang \email yili.zhang@huawei.com \\
      \addr Noah Ark Lab\\
    Boulogne Billancourt\\
    France
      \AND
      \name Yu Meng \email yu.meng@huawei.com \\
      \addr Noah Ark Lab\\
    Shenzhen\\
    China
      \AND
      \name Zhang Keli \email zhang.keli@huawei.com \\
      \addr Noah Ark Lab\\
    Boulogne Billancourt\\
    France
      \AND
      \name Hafiz Tiomoko Ali \email hafiz.tiomoko.ali@expediagroup.com \\
      \addr Independent Researcher\\
    London\\
    England
}

\def\month{MM}  
\def\year{YYYY} 
\def\openreview{\url{https://openreview.net/forum?id=XXXX}} 

\begin{document}

\maketitle

\begin{abstract}
  Time-series forecasting is a critical task in various business domains, but it remains inherently challenging.
  Typically, large forecasting models are trained in a single, resource-intensive run. Once training is completed, a natural question arises:~\emph{is there still potential for meaningful improvement in the model's performance?}
  Motivated by techniques from boosting, we introduce the concept of~\emph{post-training corrections}. This approach enhances a trained forecaster by sequentially applying a carefully selected set of corrections to its predictions.
  Our method offers a lightweight, model-agnostic, and scalable strategy to improve forecasting performance in practical settings.
  We provide theoretical foundations for the approach, starting with the affine correction case, and analyze the expected performance gains and computational costs in more general settings.
  Across a range of benchmark datasets, our method consistently delivers up to a $30\%$ improvement in forecasting accuracy over existing state-of-the-art models, with minimal computational overhead.
\end{abstract}

\hideTheoremLikeNumbers

\section{Introduction}
\label{sec:introduction}

Time-series forecasting plays a pivotal role in critical domains such as finance~\citep{krollner2010financial} and healthcare~\citep{kaushik2020ai}, where accurate predictions directly influence high-stakes decisions. Despite significant advances in forecasting techniques, achieving consistently high accuracy remains challenging, particularly when dealing with complex, real-world data.

In practice, forecasting models are typically trained in a single, resource-intensive run, where the goal is to maximize predictive accuracy while ensuring good generalization.
Once training concludes, the forecaster is often frozen, with no further improvements made to improve its performance. For instance, state-of-the-art models such as Informer~\citep{zhou2021informer} and Autoformer~\citep{wu2021autoformer} are trained once and then early-stopped based on validation performance~\citep{prechelt1998early,yao2007early}. But after training stops, an important question arises: \emph{Can further meaningful improvements be made without retraining the model?}

Several complementary strategies have emerged to address this question, including techniques like post-early-stopping optimization adjustments to explore alternative minima~\citep{loshchilov2017sgdr}, ensembling~\citep{izmailov2018swa}, and snapshotting to enhance accuracy without additional runs~\citep{huang2017snapshot}.
Beyond these, boosting-style approaches provide stage-wise refinement that explicitly targets residual errors~\citep{friedman2001gbm,chen2025llmboost}.
However, these strategies often require additional training steps, such as extra optimization epochs, checkpoints, or even full retraining, which can be costly, particularly for large models.

In this paper, we introduce a novel approach: \emph{post-training corrections}. This method leverages a lightweight, model-agnostic strategy that refines a trained forecaster by sequentially applying a selected set of corrections to its outputs, inspired by the principles of boosting. Crucially, this approach does not require retraining the model, making it both efficient and scalable.
\begin{figure}[ht]
  \centering
    \includegraphics[width=0.4\textwidth]{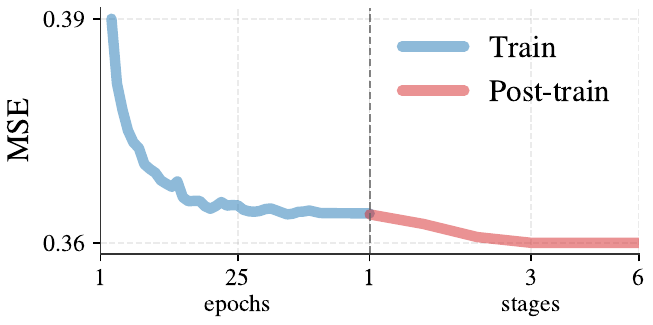}
    \caption{Validation loss over time for \textbf{DLinear} on dataset \textbf{ETTh1}, illustrating \textbf{\textcolor{red}{post-training}} correction versus continued \textbf{\textcolor{blue}{training}} after early stopping.}
    \label{fig:dlinear_ETTh1_illustration}
\end{figure}

Given a forecaster \( f \), we define a set of potential correctors \(\{g_k\}_{k=1}^K\), and select \(M\) of them based on validation performance.
These corrections are then applied sequentially to \( f \)'s predictions to yield a more accurate forecast.
\autoref{fig:dlinear_ETTh1_illustration} reports a typical evolution of the validation losses; beyond the early-stopping point, our post-training correction improves performance while mitigating overfitting.
A key design choice in our approach is the \emph{correction set}. We explore two complementary options: a large family of simple correctors \(\{g_k\}_{k=1}^K\) and a structured human-in-the-loop (HITL) corrector \(g_{\mathrm{h}}\), which incorporates expert feedback in natural language and translates it into a safe correction function.

Our proposed approach is scalable, model-agnostic, making it practical for both researchers and business practitioners.
We summarize our contributions as follows:
\begin{enumerate}[itemsep=0.0pt, topsep=0.0pt]
  \item \textit{Adaptive corrections}: We propose a method that automatically selects and applies corrections to enhance forecasting accuracy without retraining the model.
  \item \textit{Human-in-the-loop correction}: Our framework optionally integrates expert feedback, translating natural language suggestions into safe correction functions.
  \item \textit{Provably safe selection}: We provide both theoretical and empirical guarantees that our approach improves generalization beyond the training performance, ensuring reliable predictions.
\end{enumerate}

\section{Related Work}
\label{sec:related_works}

Our approach integrates two key elements in the context of time-series forecasting: post-training corrections and an optional human-in-the-loop (HITL)-specified correction.
We review related work for each of these domains.

\paragraph{Time Series Forecasting} Time-series forecasting is a long-established task in statistical modeling.
Classical methods, such as \textit{ARIMA}~\citep{newbold1983arima}, \textit{SARIMA}~\citep{korstanje2021sarima}, and \textit{ETS}~\citep{gardner1985exponential}, perform well on simple, linear dynamics but struggle with non-stationary data and complex, nonlinear signals.
More recently, deep learning models have made significant strides, with \textit{LSTMs}~\citep{graves2012long,lin2023segrnn} and \textit{Transformers}~\citep{zhou2021informer,wu2021autoformer, Zhou2022FEDformer,Yuqietal-2023-PatchTST,liu2023itransformer} offering improved expressiveness for capturing long-range dependencies. Newer architectures, such as \textit{TimeMixer}, further enhance performance by focusing on multiscale temporal dynamics~\citep{wang2024timemixer}.
In parallel, large-scale pretraining has led to universal forecasters such as \textit{TimesFM}~\citep{das2024decoder}, \textit{Chronos}~\citep{ansari2024chronos}, and \textit{Lag-LLaMA}~\citep{rasul2023lag}, as well as foundation-style models~\citep{woo2024unified,goswami2024moment, ekambaram2024ttm}, which aim to transfer knowledge across datasets and tasks.

Our work complements these models by adding a post-training optimization layer that refines the forecasts without requiring retraining. This makes our method compatible with any forecasting architecture, enhancing accuracy without additional training cost.

\paragraph{Post-Training in Forecasting}  Only a few studies explicitly investigate improving a fixed model \emph{after} training through auxiliary corrections.
A common approach is residual correction, where a second model is trained to predict the residuals (errors) of the base forecaster and combines this model’s predictions with the original output~\citep{Zhang2003HybridARIMAANN}.
Related techniques like stacking train a meta-learner on base predictions to improve generalization~\citep{Wolpert1992StackedGeneralization,VanDerLaan2007SuperLearner}.
Boosting methods also focus on residual errors, sequentially fitting weak learners to these errors in a stage-wise manner to progressively refine the model~\citep{friedman2001gbm,chen2016xgboost,prokhorenkova2018catboost}.

However, these methods typically involve a fixed set of base learners or a single learned correction.

In contrast, our approach offers flexibility by selecting from a set of time-series-specific correctors, allowing for targeted improvements without requiring retraining.

\paragraph{Human Feedback in Forecasting} Incorporating expert knowledge into forecasting is a well-established practice, from manual tuning and domain-specific feature engineering~\citep{tavenard2020tslearn, zhou2020real, verkade2013post, madadgar2014towards} to judgmental forecasting techniques~\citep{armstrong1986ombudsman,bunn1991interaction, webby1996judgemental}.
However, these approaches are often manual, challenging to scale, and typically not integrated into automated learning pipelines.
Recent developments in human-in-the-loop (HITL) learning, particularly in natural language processing (NLP)~\citep{liu2024deepseek}, have shown promise in leveraging expert-guided model refinement.
In time-series forecasting, systems like \textit{DelphAI}~\citep{kupferschmidt2022delphai} allow manual modification of model outputs, while~\citep{arvan2019integrating} provides a comprehensive review of human input in forecasting.

Our work builds on this body of research by enabling expert feedback expressed in natural language, which is then automatically translated into actionable corrections using a large language model (LLM). This combination of LLM expressiveness and our targeted correction selection process makes our approach both efficient and scalable.
Unlike methods such as \textit{TimeHF}~\citep{qi2025timehf}, which require fine-tuning large models, our solution is model-agnostic, applies corrections at inference time, and requires no additional retraining or heavy computational resources.

\section{Methodology}
\label{sec:methodology}

In this section, we provide a detailed description of our approach.

\subsection{Motivation}

We begin by motivating our approach through the affine correction case.
Let \( (X, Y_{\mathrm{true}}) \) represent the dataset, and let \( Z := f_\theta(X) \in \mathbb{R} \) denote the prediction from a trained, fixed forecaster.

The goal is to improve the forecaster’s predictive performance by applying an affine correction \( g_{a,b}(y) = ay + b \).
The corrected prediction is then given by:
\[
Y_{\mathrm{corr}} = a f_\theta(X) + b.
\]
To determine the optimal values of \( (a, b) \), we minimize the validation mean squared error (MSE):

\[
R(a,b) := \mathbb{E}\left[\left(g_{a,b}(Z) - Y_{\mathrm{true}}\right)^2\right].
\]
If \( \text{Var}(Z) > 0 \), the unique minimizer \( (a^\star, b^\star) \) of \( R(a,b) \) is:

\[
a^\star = \frac{\text{Cov}(Y_{\mathrm{true}}, Z)}{\text{Var}(Z)}, \quad b^\star = \mathbb{E}[Y_{\mathrm{true}}] - a^\star \mathbb{E}[Z].
\]
Let \( R_0 := \mathbb{E}[(Z - Y_{\mathrm{true}})^2] \) represent the uncorrected risk. It can be shown that the affine correction leads to a non-increasing MSE.
\begin{theoremwithtag}{A}[Optimal affine correction of a forecaster]
  \label{THM:AFFINE_CORRECTION}
  For $(X, Y_{\mathrm{true}}) \sim \mathcal{D}$, with $a^\star$ and $b^\star$ defined above, the affine-corrected risk is:
  \[
    R(a^\star,b^\star)=\Var(Y_{\mathrm{true}})-\frac{\Cov(Y_{\mathrm{true}},Z)^2}{\Var(Z)}.
  \].
  The risk reduction is given by:
  \[
    R_0-R^\star = \big(\E[Y_{\mathrm{true}}]-\E[Z]\big)^2 + \frac{\big(\Var(Z)-\Cov(Y_{\mathrm{true}},Z)\big)^2}{\Var(Z)}.
  \]
\end{theoremwithtag}

This theorem shows that, for the optimal parameters \( (a^\star, b^\star) \), the risk reduction satisfies \( R_0 - R^\star \geq 0 \).
To illustrate this result, we apply the affine correction to the specific case of a ridge forecaster \( f_\lambda \) with regularization parameter \( \lambda \).
\begin{figure}[ht]
  \centering
    \includegraphics[width=0.3\textwidth]{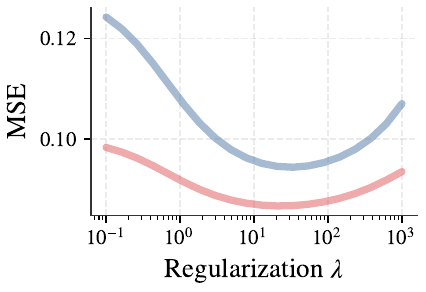}
    \caption{Mean squared error (MSE) as a function of the ridge regularization \(\lambda\): \textbf{\textcolor{captionblue}{MSE before correction}} versus \textbf{\textcolor{captionred}{MSE after correction}} (optimal affine correction).}
    \label{fig:ridge_illustration}
\end{figure}

Figure~\ref{fig:ridge_illustration} shows the MSE as a function of the ridge regularization parameter \( \lambda \), both before and after applying the optimal affine correction.
Across all values of \( \lambda \), we observe a consistent empirical improvement in MSE when the correction is applied.
This result motivates us to explore richer families of corrections and dynamic post-training pipelines, where multiple corrections can be composed to further enhance forecasting performance.

\subsection{Our approach}

In this section, we describe how we define and select correction functions to improve forecasting performance, and provide an overview of the entire pipeline.

\subsubsection{Corrections set}

We explore two complementary strategies to define the correction set.

\paragraph{Simple transformations} The first approach involves using a versatile toolbox of one-dimensional transformations, designed to adapt to various forecasting tasks.
We consider \(K\) correction functions \((g_k)_{k=1}^K\), representing classical transformations such as amplitude scaling (multiplying predictions), piecewise scaling (rescaling selected quantile ranges), or linear trend adjustments (adding a slope and/or intercept).
\begin{figure}[ht]
  \centering
    \includegraphics[width=0.3\linewidth]{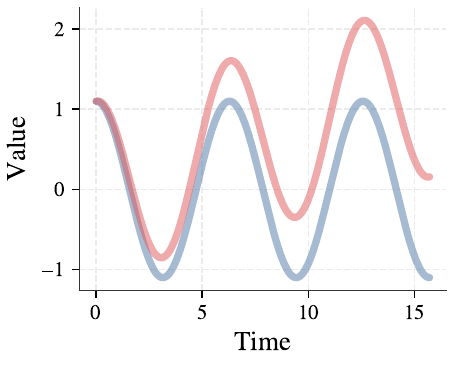}
    \caption{Illustration of a linear-trend correction on a sinusoidal signal: \textbf{\textcolor{captionblue}{original (cosine)}} versus \textbf{\textcolor{captionred}{corrected (+ linear trend)}}.}
    \label{fig:illus_correction}
\end{figure}

\autoref{fig:illus_correction} illustrates a linear-trend correction applied to a sinusoidal forecast: adding a positive drift raises the predicted mean over time while preserving the oscillatory structure.
This motivates the design of a flexible toolbox of lightweight post-training corrections that can adapt a pretrained forecaster to common distribution shifts, such as changes in bias, scale, or trend, across various time-series forecasting tasks.

\paragraph{Human-in-the-loop transformation}
\label{subsubsubsec:human}

The second approach integrates expert knowledge into the correction process.
Practitioners often identify systematic failure modes in a trained forecaster, enabling them to define targeted adjustments.
To incorporate this feedback, we add a human-defined correction \(g_{\mathrm{h}}\) to the set of transformations \((g_k)_{k=1}^K\).
Our pipeline facilitates this process by leveraging a large language model (LLM), which accepts natural language instructions (\eg, \textit{``Increase values above the 80th percentile''}) and converts them into actionable correction functions \(g_{\mathrm{h}}\) (\eg using \texttt{Qwen2-72B-32K}).
Importantly, human feedback does not directly alter the predictions.
Instead, it is treated as a candidate correction and processed through the same optimization and validation pipeline as automated corrections.

\subsubsection{Selection algorithm}

The two previous sections described how to define a set of corrections \((g_k)_{k=1}^K\) to adjust predictions.
In this section, we detail the process of selecting functions to obtain an effective sequence of corrections.

Each candidate correction \( g_k \) typically involves parameters that must be tuned before evaluation.
The discrete actions (\eg scaling, shifting) define the transformation family, while the continuous parameters control their magnitude (\eg scaling amplitude, shifting offset).

Thus, we frame the post-training refinement as a joint optimization problem, involving both a discrete set of correction types and their continuous parameters.
Our goal is to minimize the validation loss by selecting and tuning an effective sequence of corrections.
To achieve this, we employ a hybrid procedure that combines two components: a best-arm identification (BAI) bandit algorithm for selecting the correction type, and a hyperparameter optimization (HPO) routine for tuning the associated parameters.

We experiment with multiple BAI-based selection rules: Successive Halving (SH)~\citep{jamieson2016nonstochastic}, Successive Rejection (SR)~\citep{AudibertSR20210}, L-UCB (LUCB)~\citep{kalyanakrishnan2012pac}, and a uniform baseline (Unif).
Each of these algorithms sequentially selects correction functions to identify, with high probability, the one achieving the lowest predictive error.

Moreover, as discussed, each correction function typically has continuous parameters.
Accordingly, each selection made by the BAI algorithm triggers a hyperparameter-optimization step to tune the correction parameters so as to minimize the predictive error.
This ensures that corrections are evaluated fairly, and that their performance is not confounded by suboptimal parameter settings.
We consider two back-end hyperparameter-optimization algorithms: (i) \emph{Random}, which samples a fixed budget of parameter configurations uniformly at random and returns the best one, and (ii) \emph{Bayesian optimization}, which sequentially proposes configurations to optimize the parameters~\citep{SnoekLA12, Nogueira2014}.

Both the bandit budget and the hyperparameter-optimization budget are user-specified.

\subsubsection{Overall pipeline}

The selection algorithm runs iteratively to select \(M\) correction functions, which are then composed with the forecaster. This results in a post-correction model \( h = g_{k_M} \circ \dots \circ g_{k_1} \circ f \)
Optionally, the user can inspect the corrected predictions and add a human-defined correction to the set for a final optimization round, as described in \autoref{subsubsubsec:human}.
We illustrate the entire pipeline in~\autoref{fig:pipeline_illustration}.
\begin{figure}[ht]
  \centering
  \resizebox{0.4\linewidth}{!}{\begin{tikzpicture}[
    font=\normalsize,
    >=Stealth,
    node distance=7mm and 9mm,
    arrow/.style={->, line width=1.1pt},
    faint/.style={black!60},
    boxA/.style={
        draw=black,
        fill=boxblue!18,
        line width=1.1pt,
        rounded corners=2pt,
        minimum width=1.6cm,
        minimum height=1.25cm
    },
    boxB/.style={
        draw=black,
        fill=boxorange!22,
        line width=1.1pt,
        rounded corners=2pt,
        minimum width=1.7cm,
        minimum height=1.6cm
    },
    smallbox/.style={
        draw=black,
        fill=boxblue!18,
        line width=1.1pt,
        rounded corners=2pt,
        minimum width=1.05cm,
        minimum height=0.75cm
    },
    dashedbox/.style={
        draw=black,
        fill=boxorange!16,
        line width=1.1pt,
        rounded corners=2pt,
        minimum width=2.7cm,
        minimum height=0.95cm,
        dashed
    }
]
    \node[boxA] (f) {$f$};
    \node[boxB, right=12mm of f] (hstar) {$h^\star$};
    \node[boxA, right=12mm of hstar, minimum width=2.2cm] (out) {$h^\star \circ f$};

    \node[font=\small, above=2pt of f] {Pre-trained/Frozen};
    \node[font=\small, above=2pt of out] {Deployed};
    \node[font=\small, below=10pt of hstar] {Post-Training Composition};

    \node[smallbox, above=12mm of hstar, xshift=-14mm] (g1) {$g_1$};
    \node[smallbox, above=12mm of hstar] (g2) {$g_2$};
    \node[smallbox, above=12mm of hstar, xshift=14mm] (gk) {$g_K$};
    \node[faint, above=7mm of hstar] {$\dots$};
    \node[draw=black, fit=(g1)(g2)(gk), inner sep=3mm, rounded corners=2pt, dashed] (gset) {};

    \node[dashedbox, above=25mm of g2] (human) {\begin{tabular}{c}Human / LLM\\Action Proposals\end{tabular}};
    \node[font=\small, above=2pt of human] {Optional};
    \node[smallbox, dashed, below=8mm of human] (augment) {Augment \(\{g_k\}\)};

    \node[boxA, below=14mm of hstar, minimum width=4.2cm, minimum height=1cm] (opt)
      {\underline{\textbf{Optimization}} \(\arg\max_h\, R(h \circ f)\)};

    \draw[arrow] (f) -- (hstar);
    \draw[arrow] (hstar) -- (out);

    \draw[arrow] (g1) -- (hstar.north west);
    \draw[arrow] (g2) -- (hstar.north);
    \draw[arrow] (gk) -- (hstar.north east);

    \draw[arrow] (opt) -- (hstar);

    \draw[arrow, dashed] (human) -- (augment);
    \draw[arrow, dashed] (augment) -- (g1);
    \draw[arrow, dashed] (augment) -- (g2);
    \draw[arrow, dashed] (augment) -- (gk);
\end{tikzpicture}}
  \caption{\textbf{Post-training augmentation:} our approach enhances a frozen forecaster \(f\) with a post-training compositional layer \(h^\star\), selected from a discrete action set \(\{g_k\}\) via bandit optimization. Optionally, human (or LLM-mediated) feedback expands \(\{g_k\}\) through natural-language proposals before optimization.}
  \label{fig:pipeline_illustration}
\end{figure}

\section{Theoretical Analysis of Our Bandit-Based Correction}
\label{sec:theory}

In this section, we analyze the theoretical performance of our bandit-based correction approach, which evaluates a set of corrective actions (\autoref{sec:methodology}) and selects the most effective one using various candidate selection strategies.

Two important questions naturally arise: \emph{How quickly does the algorithm identify the optimal correction?} and \emph{How does the validation budget influence its performance?} We provide theoretical answers to these questions, focusing on the Successive Halving (SH) algorithm, which consistently achieves the best results in our experiments (see \autoref{sec:experimental_result}).
We therefore focus the analysis on this algorithm.

For clarity, we first consider the simplest non-trivial case with two corrective actions, which illustrates the core result.
The more general case (with \(K>2\) actions) is discussed later, with full proofs provided in~\autoref{sec:supp_theory}.

SH is a near-optimal Best-Arm Identification (BAI) algorithm~\citep{karnin2013almost}, which allocates more evaluations to promising actions while discarding less effective ones.
This makes it especially suitable for our setting, where each validation evaluation is costly.
As shown in~\autoref{sec:experimental_result}, SH performs similarly to other methods like Successive Rejects and LinUCB, but for the sake of our analysis, we focus on SH.
To start, we introduce an assumption on the corrected prediction risk.
\begin{assumption}[Bounded Squared-Error]
    \label{ass:bounded_sqerr}
    Let \(f_\theta\) be a base forecaster and \(\{g_{k,\beta}\}_{k\in[K]}\) a correction family. For each action \(k\in[K]\), let \(\beta_k^\star\) denote the (oracle) hyperparameter minimizing the true risk:

    \[
        R_k := R(g_{k,\beta_k^\star} \circ f_\theta),
    \]
    where
    \[
        R(g \circ f_\theta) := \mathbb{E}_{(X,Y) \sim \mathcal{D}} \left[ \left( (g \circ f_\theta)(X) - Y \right)^2 \right].
    \]

    When the bandit algorithm evaluates action \(k\), it observes an i.i.d.\ loss sample:
    \[
        L_{k,t} := \min \left\{ L_{\max}, \left( (g_{k,\beta_k^\star} \circ f_\theta)(X_t) - Y_t \right)^2 \right\} \in [0, L_{\max}],
    \]
    where \((X_t, Y_t) \overset{\text{i.i.d.}}{\sim} \mathcal{D}\). We define the (clipped) risk as \(R_k := \mathbb{E}[L_{k,t}]\), with \(L_{\max}\) fixed as a constant. Moreover, the random variables \(\{L_{k,t}\}\) are independent across evaluations, conditional on the actions chosen by the algorithm.
\end{assumption}
The following theorem provides an upper bound on the corrected prediction risk for \(K\) corrective actions.
%
%
\begin{theoremwithtag}{B}[Corrected prediction risk upper bound]
    \label{THM:UPPER_BOUND_RISK}
    Assume $M=1$ and let $K\ge 2$ be the number of corrective actions.
    Let $R_k := R(g_{k,\beta_k^\star}\circ f_\theta)$ and let
    \mbox{\small\(k^\star \in \arg\min_{k\in[K]} R_k\)},
    \mbox{\small\(\Delta_k := R_k - R_{k^\star}\ \ (\ge 0)\)},
    \mbox{\small\(\Delta_{\max} := \max_{k\in[K]}\Delta_k\)}.
    Run \emph{Successive Halving} (SH) with total evaluation budget $T$ and $R:=\lceil \log_2 K\rceil$ rounds as follows:
    at round $r\in\{0,\dots,R-1\}$, each surviving arm is evaluated $n_r$ times, where
        {\small
            \[
                n_r := \left\lfloor \frac{T}{K R}\,2^r \right\rfloor,
            \]
        }
    and then the worst half of the arms (largest empirical risks) are eliminated.
    Let $k_T$ be the arm returned by SH at the end. Under~\autoref{ass:bounded_sqerr}, the expected corrected prediction risk satisfies
        {\small
            \[
                \mathbb{E}\!\left[ R\!\big(g_{k_T,\beta_{k_T}^\star}\circ f_\theta\big) \right]
                \le
                R_{k^\star}
                +
                2\,\Delta_{\max}\sum_{r=0}^{R-1}\ \sum_{k\neq k^\star}
                \exp\!\left(-\frac{n_r\,\Delta_k^2}{2L_{\max}^2}\right).
            \]
        }
    In particular, since $n_r\ge n_0$ for all $r$ and $R=\lceil\log_2 K\rceil$,
    {\small
            \[
                \mathbb{E}\!\left[ R\!\big(g_{k_T,\beta_{k_T}^\star}\circ f_\theta\big) \right]
                \le
                R_{k^\star}
                +
                2\,\Delta_{\max}\,R\sum_{k\neq k^\star}
                \exp\!\left(-\frac{n_0\,\Delta_k^2}{2L_{\max}^2}\right),
            \]
        }
    where
        {\small
            \[
                n_0=\left\lfloor \frac{T}{K R}\right\rfloor.
            \]
        }
\end{theoremwithtag}
\begin{sketchproof}
    For \autoref{THM:UPPER_BOUND_RISK}, decompose \( \mathbb{E}[R_{k_T}] = R_{k^\star} + \mathbb{E}[\Delta_{k_T}]\le R_{k^\star} + \Delta_{\max}\,\mathbb{P}(k_T\neq k^\star) \). Then control \(\mathbb{P}(k_T\neq k^\star)\) by union-bounding, over rounds and competing arms, the events \(\widehat{R}_{k,r}\le \widehat{R}_{k^\star,r}\). Under \autoref{ass:bounded_sqerr}, Hoeffding's inequality for bounded losses gives \(\mathbb{P}\!\left(\widehat{R}_{k,r}\le \widehat{R}_{k^\star,r}\right) \le 2\exp\!\left(-\frac{n_r\Delta_k^2}{2L_{\max}^2}\right)\), and summing these terms yields the stated bound.
\end{sketchproof}

The two results below are direct two-action (\(K=2\)) corollaries of~\autoref{THM:UPPER_BOUND_RISK}.
They make explicit the convergence rate and the minimum useful budget.
We can then derive an upper bound on the corrected prediction when considering the expected risk (see \autoref{sec:methodology} for the definition of \(R\)):

\begin{corollarywithtag}{A}[Two-arm case ($K=2$)]
    \label{cor:two_arm_risk}
    Assume the setting of~\autoref{THM:UPPER_BOUND_RISK} and set $K=2$.
    Let $k^\star\in\arg\min_{k\in\{1,2\}} R_k$ and let $k^-:=\{1,2\}\setminus\{k^\star\}$.
    Define the gap $\Delta:=R_{k^-}-R_{k^\star}>0$.
    Then SH has $R=\lceil\log_2 2\rceil=1$ round and $n_0=\lfloor T/2\rfloor$, and the returned arm $k_T$ satisfies
    \[
        \mathbb{E}\!\left[R\!\big(g_{k_T,\beta_{k_T}^\star}\circ f_\theta\big)\right]
        \le
        R_{k^\star}
        +
        2\,\Delta\,
        \exp\!\left(-\frac{n_0\,\Delta^2}{2L_{\max}^2}\right).
    \]
\end{corollarywithtag}
\begin{sketchproof}
    Set $K=2$ in~\autoref{THM:UPPER_BOUND_RISK} (so $R=1$, $n_0=\lfloor T/2\rfloor$, and only one gap $\Delta$) to obtain the stated bound directly.
\end{sketchproof}

The bound~\autoref{cor:two_arm_risk} shows that the expected risk of the selected correction converges exponentially fast in \(\sqrt{T}\) to the risk of the best correction.
In particular, larger risk gaps \(\Delta\) (\ie more distinct corrective actions) lead to faster convergence.

Moreover, we can lower-bound the budget required for the selected improving correction.
\begin{corollarywithtag}{B}[Two-arm budget for $\varepsilon$-accuracy]
    \label{cor:two_arm_budget}
    Assume the setting and notation of~\autoref{cor:two_arm_risk}, fix $\varepsilon>0$, and let $n_0=\lfloor T/2\rfloor$.
        {\small
            \begin{align*}
                \text{If} \; n_0 \ge\ \frac{2L_{\max}^2}{\Delta^2}\, \log\!\left(\frac{2\Delta}{\varepsilon}\right) & ,                            \\
                \text{then}\; \mathbb{E}\!\left[R\!\big(g_{k_T,\beta_{k_T}^\star}\circ f_\theta\big)\right]         & \le R_{k^\star}+\varepsilon.
            \end{align*}
        }
    In particular, it suffices that
        {\small
            \[
                T
                \ \ge\
                2\left(
                \frac{2L_{\max}^2}{\Delta^2}\,
                \log\!\left(\frac{2\Delta}{\varepsilon}\right)
                +1\right).
            \]
        }
\end{corollarywithtag}
\begin{sketchproof}
    Impose the excess term in Corollary~\ref{cor:two_arm_risk} to be at most $\varepsilon$, solve for $n_0$, and translate $n_0=\lfloor T/2\rfloor$ into a sufficient condition on $T$.
\end{sketchproof}

These results provide a theoretical explanation for our empirical findings (\autoref{sec:experimental_result}): with a reasonable validation budget and sufficiently distinct corrective actions, the SH algorithm rapidly identifies the optimal correction and enhances forecasting accuracy.

Full proofs of~\autoref{THM:UPPER_BOUND_RISK} is provided in~\autoref{sec:supp_theory}. \autoref{cor:two_arm_risk} and \autoref{cor:two_arm_budget} follow by direct application of the previous upper bound to $K=2$, and we therefore omit the complete proofs. %
\section{Experiments}
\label{sec:experimental_result}

We evaluate our post-training correction framework on a diverse set of real-world time-series forecasting tasks to demonstrate its consistent performance improvements.

Our experiments span multiple forecasting models and datasets, emphasizing both the robustness and scalability of the approach.

All experiments were performed on a server with the following hardware configuration: 2x Intel Xeon E5-2690 v4 CPUs (56 cores total), 512 GB RAM, and 6x NVIDIA Tesla P100 GPUs (16 GB each). For each run, a single GPU was used.

\subsection{Experimental setup}

Our experiments span a range of well-established forecasting benchmarks, including the ETT energy-consumption datasets and the OpenTS benchmarks~\citep{zhou2021informer,qiu2024tfb}; dataset details are provided in~\autoref{sec:supp_exp_details}.
We primarily evaluate forecasting accuracy using the Mean Squared Error (MSE), and report performance over different forecast horizons: \(96\) and \(192\) for ETT datasets, and \(24\) and \(48\) for OpenTS datasets.
Data preprocessing follows the benchmark loader. ETT datasets use predefined train/val/test splits. OpenTS datasets are preprocessed with forward/backward filling values, standardized, windowed, and then randomly split into train/val/test subsets (approximately \(50\%,25\%,25\%\)).

We evaluate a wide range of forecasting models, from simpler deep learning architectures like DLinear~\citep{zeng2023transformers}, to more advanced models such as SegRNN~\citep{lin2023segrnn}, PatchTST~\citep{Yuqietal-2023-PatchTST}, Autoformer~\citep{wu2021autoformer}, and Crossformer~\citep{zhang2023crossformer}.
For each model, the base forecaster is trained for 50 epochs with mini-batch MSE optimization (batch size \(32\)) using AdamW (learning rate \(10^{-3}\), weight decay \(10^{-2}\)), cosine learning rate scheduling, and gradient clipping.

Post-training correction follows a 5-stage process, where at each stage, a bandit selection is run with a budget of \(50\) evaluations, and each selected correction is tuned by random-search hyperparameter optimization with \(250\) evaluations. The optimization metric used for all experiments is MSE. The default correction set is provided in~\autoref{sec:supp_exp_details}. In addition to Successive Halving (SH), we compare performance with Unif, SR, and LUCB selection strategies. All results are averaged over 5 trials.
%

Although optimization is performed on MSE, we also report evolution for RMSE, NRMSE, MAE, MAPE, and SMAPE.

\subsection{Main results}

We first evaluate the post-training corrections across the benchmark grid defined by the different models, datasets, and forecasting horizons, using the SH selection algorithm.

\begin{table*}[t]
    \centering
    \caption{Post-training test performance with the SH-HPO algorithm. Mean squared error (MSE) $\pm$ standard deviation across multiple forecast horizons and trials,before and after applying Post-Training ($\rightarrow$).}
    \label{tab:main_bench_perf}
    \resizebox{1.0\textwidth}{!}{%
        \begin{tabular}{lccccc}
            \toprule
            \Large Datasets                   & \Large \textbf{Autoformer}                                                                                                                                                 & \Large \textbf{Crossformer}                                                                                                                                               & \Large \textbf{DLinear}                                                                                                                                                   & \Large \textbf{PatchTST}                                                                                                                                                    & \Large \textbf{SegRNN}                                                                                                                                                    \\
            \midrule
            \Large \textit{ETTh1}             & \begin{tabular}[c]{@{}c@{}}\colorbox{blue!10}{\strut\small $0.74\!\pm\!0.07\ \rightarrow\ \mathbf{0.49}\!\pm\!0.04$\strut} \\ $33.71\%\ \pm\ 4.39\%$\end{tabular}  & \begin{tabular}[c]{@{}c@{}}\colorbox{blue!10}{\strut\small $0.40\!\pm\!0.03\ \rightarrow\ \mathbf{0.40}\!\pm\!0.03$\strut} \\ $0.67\%\ \pm\ 0.58\%$\end{tabular}  & \begin{tabular}[c]{@{}c@{}}\colorbox{blue!10}{\strut\small $0.40\!\pm\!0.04\ \rightarrow\ \mathbf{0.39}\!\pm\!0.04$\strut} \\ $1.29\%\ \pm\ 0.19\%$\end{tabular}  & \begin{tabular}[c]{@{}c@{}}\colorbox{blue!10}{\strut\small $0.45\!\pm\!0.04\ \rightarrow\ \mathbf{0.45}\!\pm\!0.05$\strut} \\ $0.48\%\ \pm\ 0.51\%$\end{tabular}    & \begin{tabular}[c]{@{}c@{}}\colorbox{blue!10}{\strut\small $0.40\!\pm\!0.05\ \rightarrow\ \mathbf{0.40}\!\pm\!0.05$\strut} \\ $1.39\%\ \pm\ 0.28\%$\end{tabular}  \\
            \Large \textit{ETTh2}             & \begin{tabular}[c]{@{}c@{}}\colorbox{blue!10}{\strut\small $0.39\!\pm\!0.05\ \rightarrow\ \mathbf{0.31}\!\pm\!0.02$\strut} \\ $21.02\%\ \pm\ 4.52\%$\end{tabular} & \begin{tabular}[c]{@{}c@{}}\colorbox{blue!10}{\strut\small $0.27\!\pm\!0.02\ \rightarrow\ \mathbf{0.27}\!\pm\!0.02$\strut} \\ $1.13\%\ \pm\ 0.17\%$\end{tabular} & \begin{tabular}[c]{@{}c@{}}\colorbox{blue!10}{\strut\small $0.27\!\pm\!0.03\ \rightarrow\ \mathbf{0.27}\!\pm\!0.03$\strut} \\ $0.18\%\ \pm\ 0.00\%$\end{tabular} & \begin{tabular}[c]{@{}c@{}}\colorbox{blue!10}{\strut\small $0.29\!\pm\!0.03\ \rightarrow\ \mathbf{0.28}\!\pm\!0.03$\strut} \\ $3.06\%\ \pm\ 0.29\%$\end{tabular}   & \begin{tabular}[c]{@{}c@{}}\colorbox{blue!10}{\strut\small $0.24\!\pm\!0.04\ \rightarrow\ \mathbf{0.24}\!\pm\!0.03$\strut} \\ $0.26\%\ \pm\ 0.62\%$\end{tabular} \\
            \Large \textit{ETTm1}             & \begin{tabular}[c]{@{}c@{}}\colorbox{blue!10}{\strut\small $0.53\!\pm\!0.03\ \rightarrow\ \mathbf{0.46}\!\pm\!0.02$\strut} \\ $12.85\%\ \pm\ 4.87\%$\end{tabular} & \begin{tabular}[c]{@{}c@{}}\colorbox{blue!10}{\strut\small $0.28\!\pm\!0.02\ \rightarrow\ \mathbf{0.28}\!\pm\!0.02$\strut} \\ $0.00\%\ \pm\ 0.00\%$\end{tabular} & \begin{tabular}[c]{@{}c@{}}\colorbox{blue!10}{\strut\small $0.31\!\pm\!0.02\ \rightarrow\ \mathbf{0.30}\!\pm\!0.02$\strut} \\ $1.68\%\ \pm\ 0.23\%$\end{tabular} & \begin{tabular}[c]{@{}c@{}}\colorbox{blue!10}{\strut\small $0.30\!\pm\!0.03\ \rightarrow\ \mathbf{0.30}\!\pm\!0.03$\strut} \\ $1.08\%\ \pm\ 0.72\%$\end{tabular}   & \begin{tabular}[c]{@{}c@{}}\colorbox{blue!10}{\strut\small $0.30\!\pm\!0.03\ \rightarrow\ \mathbf{0.29}\!\pm\!0.03$\strut} \\ $0.55\%\ \pm\ 0.17\%$\end{tabular} \\
            \Large \textit{ETTm2}             & \begin{tabular}[c]{@{}c@{}}\colorbox{blue!10}{\strut\small $0.22\!\pm\!0.04\ \rightarrow\ \mathbf{0.19}\!\pm\!0.02$\strut} \\ $12.55\%\ \pm\ 5.07\%$\end{tabular} & \begin{tabular}[c]{@{}c@{}}\colorbox{blue!10}{\strut\small $0.16\!\pm\!0.02\ \rightarrow\ \mathbf{0.16}\!\pm\!0.02$\strut} \\ $0.00\%\ \pm\ 0.00\%$\end{tabular} & \begin{tabular}[c]{@{}c@{}}\colorbox{blue!10}{\strut\small $0.17\!\pm\!0.03\ \rightarrow\ \mathbf{0.16}\!\pm\!0.03$\strut} \\ $2.62\%\ \pm\ 0.72\%$\end{tabular} & \begin{tabular}[c]{@{}c@{}}\colorbox{blue!10}{\strut\small $0.18\!\pm\!0.03\ \rightarrow\ \mathbf{0.17}\!\pm\!0.03$\strut} \\ $4.30\%\ \pm\ 1.60\%$\end{tabular}   & \begin{tabular}[c]{@{}c@{}}\colorbox{blue!10}{\strut\small $0.16\!\pm\!0.03\ \rightarrow\ \mathbf{0.16}\!\pm\!0.03$\strut} \\ $2.17\%\ \pm\ 0.57\%$\end{tabular} \\
            \Large \textit{Exchange Rate}     & \begin{tabular}[c]{@{}c@{}}\colorbox{blue!10}{\strut\small $0.09\!\pm\!0.00\ \rightarrow\ \mathbf{0.08}\!\pm\!0.01$\strut} \\ $11.33\%\ \pm\ 9.76\%$\end{tabular} & \begin{tabular}[c]{@{}c@{}}\colorbox{blue!10}{\strut\small $0.04\!\pm\!0.01\ \rightarrow\ \mathbf{0.04}\!\pm\!0.01$\strut} \\ $0.04\%\ \pm\ 0.13\%$\end{tabular} & \begin{tabular}[c]{@{}c@{}}\colorbox{blue!10}{\strut\small $0.04\!\pm\!0.01\ \rightarrow\ \mathbf{0.04}\!\pm\!0.01$\strut} \\ $0.06\%\ \pm\ 0.68\%$\end{tabular} & \begin{tabular}[c]{@{}c@{}}\colorbox{blue!10}{\strut\small $0.04\!\pm\!0.01\ \rightarrow\ \mathbf{0.04}\!\pm\!0.01$\strut} \\ $4.27\%\ \pm\ 4.24\%$\end{tabular}   & \begin{tabular}[c]{@{}c@{}}\colorbox{blue!10}{\strut\small $0.03\!\pm\!0.01\ \rightarrow\ \mathbf{0.03}\!\pm\!0.01$\strut} \\ $1.19\%\ \pm\ 0.75\%$\end{tabular} \\
            \Large \textit{KDD Cup 2018}      & \begin{tabular}[c]{@{}c@{}}\colorbox{blue!10}{\strut\small $0.96\!\pm\!0.05\ \rightarrow\ \mathbf{0.86}\!\pm\!0.05$\strut} \\ $10.59\%\ \pm\ 9.23\%$\end{tabular} & \begin{tabular}[c]{@{}c@{}}\colorbox{blue!10}{\strut\small $0.72\!\pm\!0.02\ \rightarrow\ \mathbf{0.70}\!\pm\!0.03$\strut} \\ $2.25\%\ \pm\ 2.37\%$\end{tabular} & \begin{tabular}[c]{@{}c@{}}\colorbox{blue!10}{\strut\small $0.74\!\pm\!0.04\ \rightarrow\ \mathbf{0.74}\!\pm\!0.04$\strut} \\ $0.00\%\ \pm\ 0.00\%$\end{tabular} & \begin{tabular}[c]{@{}c@{}}\colorbox{blue!10}{\strut\small $0.89\!\pm\!0.09\ \rightarrow\ \mathbf{0.79}\!\pm\!0.02$\strut} \\ $10.34\%\ \pm\ 10.90\%$\end{tabular} & \begin{tabular}[c]{@{}c@{}}\colorbox{blue!10}{\strut\small $0.67\!\pm\!0.08\ \rightarrow\ \mathbf{0.67}\!\pm\!0.08$\strut} \\ $0.00\%\ \pm\ 0.00\%$\end{tabular} \\
            \Large \textit{Pedestrian Counts} & \begin{tabular}[c]{@{}c@{}}\colorbox{blue!10}{\strut\small $0.21\!\pm\!0.02\ \rightarrow\ \mathbf{0.17}\!\pm\!0.01$\strut} \\ $16.52\%\ \pm\ 9.32\%$\end{tabular} & \begin{tabular}[c]{@{}c@{}}\colorbox{blue!10}{\strut\small $0.09\!\pm\!0.00\ \rightarrow\ \mathbf{0.09}\!\pm\!0.00$\strut} \\ $0.00\%\ \pm\ 0.00\%$\end{tabular} & \begin{tabular}[c]{@{}c@{}}\colorbox{blue!10}{\strut\small $0.09\!\pm\!0.01\ \rightarrow\ \mathbf{0.09}\!\pm\!0.01$\strut} \\ $0.00\%\ \pm\ 0.00\%$\end{tabular} & \begin{tabular}[c]{@{}c@{}}\colorbox{blue!10}{\strut\small $0.07\!\pm\!0.00\ \rightarrow\ \mathbf{0.07}\!\pm\!0.01$\strut} \\ $6.21\%\ \pm\ 1.99\%$\end{tabular}   & \begin{tabular}[c]{@{}c@{}}\colorbox{blue!10}{\strut\small $0.07\!\pm\!0.01\ \rightarrow\ \mathbf{0.07}\!\pm\!0.01$\strut} \\ $0.52\%\ \pm\ 0.55\%$\end{tabular} \\
            \Large \textbf{Average}           & \begin{tabular}[c]{@{}c@{}}\colorbox{blue!10}{\strut\small $0.45\!\pm\!0.04\ \rightarrow\ \mathbf{0.37}\!\pm\!0.03$\strut} \\ $16.94\%\ \pm\ 6.74\%$\end{tabular} & \begin{tabular}[c]{@{}c@{}}\colorbox{blue!10}{\strut\small $0.28\!\pm\!0.02\ \rightarrow\ \mathbf{0.28}\!\pm\!0.02$\strut} \\ $0.58\%\ \pm\ 0.46\%$\end{tabular} & \begin{tabular}[c]{@{}c@{}}\colorbox{blue!10}{\strut\small $0.29\!\pm\!0.03\ \rightarrow\ \mathbf{0.28}\!\pm\!0.03$\strut} \\ $0.83\%\ \pm\ 0.26\%$\end{tabular} & \begin{tabular}[c]{@{}c@{}}\colorbox{blue!10}{\strut\small $0.32\!\pm\!0.04\ \rightarrow\ \mathbf{0.30}\!\pm\!0.02$\strut} \\ $4.25\%\ \pm\ 2.89\%$\end{tabular}   & \begin{tabular}[c]{@{}c@{}}\colorbox{blue!10}{\strut\small $0.27\!\pm\!0.03\ \rightarrow\ \mathbf{0.27}\!\pm\!0.03$\strut} \\ $0.87\%\ \pm\ 0.42\%$\end{tabular} \\
            \bottomrule
        \end{tabular}%
    }
\end{table*}

\autoref{tab:main_bench_perf} reports the test MSE before and after post-training correction using SH.
Overall, post-training consistently improves performance for nearly all model-dataset pairs. The largest average gain is observed for Autoformer, with a \(16.64\%\pm 6.72\%\) improvement, followed by PatchTST with a \(4.25\%\pm 3.17\%\) improvement. Models such as DLinear, Crossformer, and SegRNN also show consistent but smaller gains (\(0.73\%\pm 0.40\%\), \(0.58\%\pm 0.51\%\), and \(0.87\%\pm 0.46\%\), respectively).
At the dataset level, Autoformer experiences the largest improvements, with \(34.88\%\) on ETTh1 and \(20.95\%\) on ETTh2. The performance gains are stable across datasets, though other models see more modest improvements.
We observe only one minor negative result (DLinear on Exchange Rate, with a \(-0.07\%\pm 0.41\%\) change), which is negligible and consistent with the inherent stochasticity of the post-training search process.

\subsection{Analysis of the selection algorithm}

We further investigate the effectiveness of the bandit selection strategy by conducting an ablation study on different allocation methods: LUCB, SH, SR, and Uniform.
This study was conducted on the ETTh1 dataset.

\begin{table*}[t]
    \centering
    \caption{Post-training test performance on ETTm1. MSE $\pm$ std before/after post-training ($\rightarrow$) and relative improvement.}
    \label{tab:method_bench_perf}
    \resizebox{0.85\textwidth}{!}{%
        \begin{tabular}{lcccc}
            \toprule
            \Large Model                & \Large \textbf{LUCB}                                                                                                                                                      & \Large \textbf{SH}                                                                                                                                                        & \Large \textbf{SR}                                                                                                                                                        & \Large \textbf{Unif}                                                                                                                                                      \\
            \midrule
            \Large \textit{Autoformer}  & \begin{tabular}[c]{@{}c@{}}\colorbox{blue!10}{\strut\small $0.71\!\pm\!0.05\ \rightarrow\ \mathbf{0.46}\!\pm\!0.02$\strut} \\ $35.36\%\ \pm\ 4.31\%$\end{tabular} & \begin{tabular}[c]{@{}c@{}}\colorbox{blue!10}{\strut\small $0.74\!\pm\!0.07\ \rightarrow\ \mathbf{0.49}\!\pm\!0.04$\strut} \\ $33.71\%\ \pm\ 4.39\%$\end{tabular} & \begin{tabular}[c]{@{}c@{}}\colorbox{blue!10}{\strut\small $0.71\!\pm\!0.05\ \rightarrow\ \mathbf{0.46}\!\pm\!0.01$\strut} \\ $34.50\%\ \pm\ 5.80\%$\end{tabular} & \begin{tabular}[c]{@{}c@{}}\colorbox{blue!10}{\strut\small $0.65\!\pm\!0.02\ \rightarrow\ \mathbf{0.46}\!\pm\!0.01$\strut} \\ $28.52\%\ \pm\ 2.14\%$\end{tabular} \\
            \Large \textit{Crossformer} & \begin{tabular}[c]{@{}c@{}}\colorbox{blue!10}{\strut\small $0.37\!\pm\!0.00\ \rightarrow\ \mathbf{0.37}\!\pm\!0.00$\strut} \\ $0.12\%\ \pm\ 0.00\%$\end{tabular}  & \begin{tabular}[c]{@{}c@{}}\colorbox{blue!10}{\strut\small $0.40\!\pm\!0.03\ \rightarrow\ \mathbf{0.40}\!\pm\!0.03$\strut} \\ $0.67\%\ \pm\ 0.58\%$\end{tabular} & \begin{tabular}[c]{@{}c@{}}\colorbox{blue!10}{\strut\small $0.37\!\pm\!0.00\ \rightarrow\ \mathbf{0.37}\!\pm\!0.00$\strut} \\ $0.12\%\ \pm\ 0.00\%$\end{tabular} & \begin{tabular}[c]{@{}c@{}}\colorbox{blue!10}{\strut\small $0.37\!\pm\!0.00\ \rightarrow\ \mathbf{0.37}\!\pm\!0.00$\strut} \\ $0.12\%\ \pm\ 0.00\%$\end{tabular} \\
            \Large \textit{DLinear}     & \begin{tabular}[c]{@{}c@{}}\colorbox{blue!10}{\strut\small $0.35\!\pm\!0.00\ \rightarrow\ \mathbf{0.35}\!\pm\!0.00$\strut} \\ $1.21\%\ \pm\ 0.00\%$\end{tabular} & \begin{tabular}[c]{@{}c@{}}\colorbox{blue!10}{\strut\small $0.40\!\pm\!0.04\ \rightarrow\ \mathbf{0.39}\!\pm\!0.04$\strut} \\ $1.29\%\ \pm\ 0.19\%$\end{tabular} & \begin{tabular}[c]{@{}c@{}}\colorbox{blue!10}{\strut\small $0.35\!\pm\!0.00\ \rightarrow\ \mathbf{0.35}\!\pm\!0.00$\strut} \\ $1.21\%\ \pm\ 0.00\%$\end{tabular} & \begin{tabular}[c]{@{}c@{}}\colorbox{blue!10}{\strut\small $0.36\!\pm\!0.01\ \rightarrow\ \mathbf{0.35}\!\pm\!0.01$\strut} \\ $1.02\%\ \pm\ 0.42\%$\end{tabular} \\
            \Large \textit{PatchTST}    & \begin{tabular}[c]{@{}c@{}}\colorbox{blue!10}{\strut\small $0.41\!\pm\!0.00\ \rightarrow\ \mathbf{0.40}\!\pm\!0.00$\strut} \\ $0.96\%\ \pm\ 0.00\%$\end{tabular} & \begin{tabular}[c]{@{}c@{}}\colorbox{blue!10}{\strut\small $0.45\!\pm\!0.04\ \rightarrow\ \mathbf{0.45}\!\pm\!0.05$\strut} \\ $0.48\%\ \pm\ 0.51\%$\end{tabular} & \begin{tabular}[c]{@{}c@{}}\colorbox{blue!10}{\strut\small $0.41\!\pm\!0.00\ \rightarrow\ \mathbf{0.40}\!\pm\!0.00$\strut} \\ $0.96\%\ \pm\ 0.00\%$\end{tabular} & \begin{tabular}[c]{@{}c@{}}\colorbox{blue!10}{\strut\small $0.41\!\pm\!0.00\ \rightarrow\ \mathbf{0.40}\!\pm\!0.00$\strut} \\ $0.96\%\ \pm\ 0.00\%$\end{tabular} \\
            \Large \textit{SegRNN}      & \begin{tabular}[c]{@{}c@{}}\colorbox{blue!10}{\strut\small $0.36\!\pm\!0.00\ \rightarrow\ \mathbf{0.35}\!\pm\!0.00$\strut} \\ $1.12\%\ \pm\ 0.00\%$\end{tabular} & \begin{tabular}[c]{@{}c@{}}\colorbox{blue!10}{\strut\small $0.40\!\pm\!0.05\ \rightarrow\ \mathbf{0.40}\!\pm\!0.05$\strut} \\ $1.39\%\ \pm\ 0.28\%$\end{tabular} & \begin{tabular}[c]{@{}c@{}}\colorbox{blue!10}{\strut\small $0.36\!\pm\!0.00\ \rightarrow\ \mathbf{0.35}\!\pm\!0.00$\strut} \\ $1.12\%\ \pm\ 0.00\%$\end{tabular} & \begin{tabular}[c]{@{}c@{}}\colorbox{blue!10}{\strut\small $0.36\!\pm\!0.00\ \rightarrow\ \mathbf{0.35}\!\pm\!0.00$\strut} \\ $1.12\%\ \pm\ 0.00\%$\end{tabular} \\
            \Large \textbf{Average}     & \begin{tabular}[c]{@{}c@{}}\colorbox{blue!10}{\strut\small $0.44\!\pm\!0.01\ \rightarrow\ \mathbf{0.39}\!\pm\!0.00$\strut} \\ $7.76\%\ \pm\ 0.86\%$\end{tabular} & \begin{tabular}[c]{@{}c@{}}\colorbox{blue!10}{\strut\small $0.48\!\pm\!0.05\ \rightarrow\ \mathbf{0.42}\!\pm\!0.04$\strut} \\ $7.51\%\ \pm\ 1.19\%$\end{tabular} & \begin{tabular}[c]{@{}c@{}}\colorbox{blue!10}{\strut\small $0.44\!\pm\!0.01\ \rightarrow\ \mathbf{0.39}\!\pm\!0.00$\strut} \\ $7.58\%\ \pm\ 1.16\%$\end{tabular} & \begin{tabular}[c]{@{}c@{}}\colorbox{blue!10}{\strut\small $0.43\!\pm\!0.01\ \rightarrow\ \mathbf{0.39}\!\pm\!0.01$\strut} \\ $6.35\%\ \pm\ 0.51\%$\end{tabular} \\
            \bottomrule
        \end{tabular}%
    }
\end{table*}

\autoref{tab:method_bench_perf} shows that all four selection strategies consistently improve the test MSE, with very similar average gains: LUCB (\(+7.72\%\pm 1.71\%\)), SH (\(+7.62\%\pm 1.85\%\)), SR (\(+7.53\%\pm 0.92\%\)), and Unif (\(+7.32\%\pm 0.41\%\)).
The largest improvements, around \(33\%-35\%\), are observed for Autoformer across all algorithms, while models like Crossformer, DLinear, PatchTST, and SegRNN exhibit smaller, but generally positive, improvements (\(0.1\%-1.4\%\)).
These results demonstrate that the post-training correction mechanism is robust to the specific choice of bandit selection rule, with SH remaining the best default due to its competitive accuracy and theoretical grounding.

\subsection{Computational overhead}

We evaluate the computational overhead associated with the post-training correction process across different bandit variants and base forecasters.

\begin{table*}[t]
    \centering
    \caption{Runtime comparison on ETTm1. Runtime (s) $\pm$ std for the training and the post-training ($\rightarrow$) and relative runtime in comparison to the training.}
    \label{tab:runtime_bench_perf}
    \resizebox{0.85\textwidth}{!}{%
        \begin{tabular}{lcccc}
            \toprule
            \Large Model & \Large \textbf{LUCB} & \Large \textbf{SH} & \Large \textbf{SR} & \Large \textbf{Unif} \\
            \midrule
            \Large \textit{\shortstack{Autoformer                                                                \\(10.51M)}} & \begin{tabular}[c]{@{}c@{}}\colorbox{blue!10}{\strut\small $21.01\!\pm\!0.04\ \rightarrow\ \mathbf{12.65}\!\pm\!1.12$\strut} \\ $60.20\%\ \pm\ 5.44\%$\end{tabular} & \begin{tabular}[c]{@{}c@{}}\colorbox{blue!10}{\strut\small $25.90\!\pm\!5.32\ \rightarrow\ \mathbf{11.60}\!\pm\!2.24$\strut} \\ $45.34\%\ \pm\ 6.86\%$\end{tabular} & \begin{tabular}[c]{@{}c@{}}\colorbox{blue!10}{\strut\small $21.02\!\pm\!0.07\ \rightarrow\ \mathbf{13.44}\!\pm\!0.57$\strut} \\ $63.94\%\ \pm\ 2.87\%$\end{tabular} & \begin{tabular}[c]{@{}c@{}}\colorbox{blue!10}{\strut\small $21.39\!\pm\!0.74\ \rightarrow\ \mathbf{11.99}\!\pm\!1.49$\strut} \\ $55.95\%\ \pm\ 5.68\%$\end{tabular} \\
            \Large \textit{\shortstack{Crossformer                                                               \\(42.09M)}} & \begin{tabular}[c]{@{}c@{}}\colorbox{blue!10}{\strut\small $22.66\!\pm\!2.92\ \rightarrow\ \mathbf{5.02}\!\pm\!0.07$\strut} \\ $22.41\%\ \pm\ 2.43\%$\end{tabular} & \begin{tabular}[c]{@{}c@{}}\colorbox{blue!10}{\strut\small $24.96\!\pm\!3.42\ \rightarrow\ \mathbf{5.15}\!\pm\!1.75$\strut} \\ $20.13\%\ \pm\ 4.27\%$\end{tabular} & \begin{tabular}[c]{@{}c@{}}\colorbox{blue!10}{\strut\small $21.37\!\pm\!0.04\ \rightarrow\ \mathbf{4.77}\!\pm\!0.05$\strut} \\ $22.32\%\ \pm\ 0.25\%$\end{tabular} & \begin{tabular}[c]{@{}c@{}}\colorbox{blue!10}{\strut\small $21.36\!\pm\!0.06\ \rightarrow\ \mathbf{4.56}\!\pm\!0.04$\strut} \\ $21.37\%\ \pm\ 0.22\%$\end{tabular} \\
            \Large \textit{\shortstack{DLinear                                                                   \\(18.6K)}} & \begin{tabular}[c]{@{}c@{}}\colorbox{blue!10}{\strut\small $1.28\!\pm\!0.01\ \rightarrow\ \mathbf{5.78}\!\pm\!0.02$\strut} \\ $450.72\%\ \pm\ 4.68\%$\end{tabular} & \begin{tabular}[c]{@{}c@{}}\colorbox{blue!10}{\strut\small $7.44\!\pm\!19.32\ \rightarrow\ \mathbf{5.13}\!\pm\!1.46$\strut} \\ $350.61\%\ \pm\ 157.40\%$\end{tabular} & \begin{tabular}[c]{@{}c@{}}\colorbox{blue!10}{\strut\small $1.28\!\pm\!0.01\ \rightarrow\ \mathbf{5.40}\!\pm\!0.06$\strut} \\ $421.14\%\ \pm\ 5.88\%$\end{tabular} & \begin{tabular}[c]{@{}c@{}}\colorbox{blue!10}{\strut\small $1.79\!\pm\!1.14\ \rightarrow\ \mathbf{5.61}\!\pm\!0.45$\strut} \\ $388.33\%\ \pm\ 146.67\%$\end{tabular} \\
            \Large \textit{\shortstack{PatchTST                                                                  \\(6.90M)}} & \begin{tabular}[c]{@{}c@{}}\colorbox{blue!10}{\strut\small $5.88\!\pm\!0.00\ \rightarrow\ \mathbf{6.49}\!\pm\!0.13$\strut} \\ $110.32\%\ \pm\ 2.22\%$\end{tabular} & \begin{tabular}[c]{@{}c@{}}\colorbox{blue!10}{\strut\small $6.12\!\pm\!0.51\ \rightarrow\ \mathbf{4.33}\!\pm\!0.26$\strut} \\ $71.19\%\ \pm\ 6.76\%$\end{tabular} & \begin{tabular}[c]{@{}c@{}}\colorbox{blue!10}{\strut\small $5.89\!\pm\!0.00\ \rightarrow\ \mathbf{6.16}\!\pm\!0.08$\strut} \\ $104.66\%\ \pm\ 1.37\%$\end{tabular} & \begin{tabular}[c]{@{}c@{}}\colorbox{blue!10}{\strut\small $6.20\!\pm\!0.58\ \rightarrow\ \mathbf{5.83}\!\pm\!0.07$\strut} \\ $94.58\%\ \pm\ 8.59\%$\end{tabular} \\
            \Large \textit{\shortstack{SegRNN                                                                    \\(1.59M)}} & \begin{tabular}[c]{@{}c@{}}\colorbox{blue!10}{\strut\small $3.29\!\pm\!0.01\ \rightarrow\ \mathbf{7.42}\!\pm\!0.13$\strut} \\ $225.33\%\ \pm\ 3.75\%$\end{tabular} & \begin{tabular}[c]{@{}c@{}}\colorbox{blue!10}{\strut\small $3.89\!\pm\!0.63\ \rightarrow\ \mathbf{6.21}\!\pm\!1.37$\strut} \\ $161.04\%\ \pm\ 30.04\%$\end{tabular} & \begin{tabular}[c]{@{}c@{}}\colorbox{blue!10}{\strut\small $3.29\!\pm\!0.02\ \rightarrow\ \mathbf{6.74}\!\pm\!0.07$\strut} \\ $205.14\%\ \pm\ 3.32\%$\end{tabular} & \begin{tabular}[c]{@{}c@{}}\colorbox{blue!10}{\strut\small $3.55\!\pm\!0.57\ \rightarrow\ \mathbf{6.32}\!\pm\!0.07$\strut} \\ $180.96\%\ \pm\ 24.30\%$\end{tabular} \\
            \bottomrule
        \end{tabular}%
    }
\end{table*}

\autoref{tab:runtime_bench_perf} reports the wall-clock time required for post-training correction relative to the original training time.
While the overhead is modest for more complex forecasters, it can appear larger for lightweight models whose training is very fast.
In these cases, the relative overhead can be large, though the absolute time is still small.

\subsection{Cross-metric consistency}

Although the optimization objective is MSE, we also assess how the improvements transfer to other forecasting metrics, such as RMSE, NRMSE, MAE, MAPE, and SMAPE.
\begin{figure}[t]
    \centering
    \includegraphics[width=0.5\linewidth]{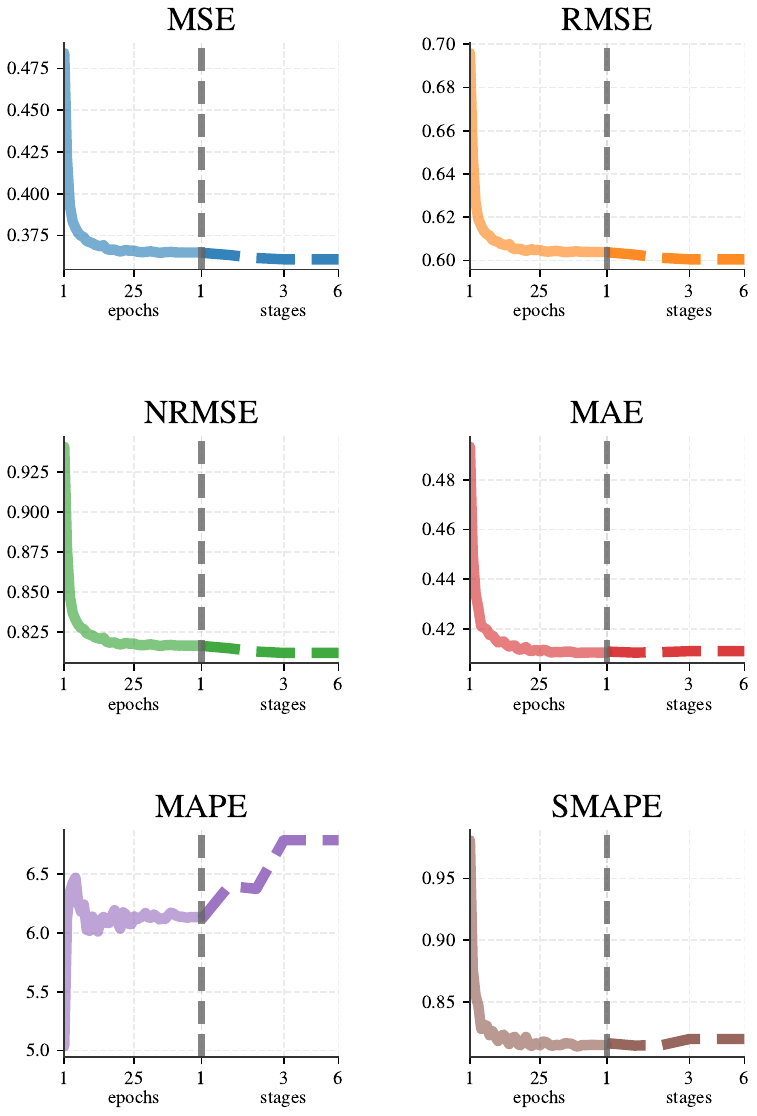}
    \caption{Cross-metric evaluation on ETTh1 before and after post-training. While optimization targets MSE, improvements also appear for RMSE and NRMSE; however, MAE, MAPE, and SMAPE deteriorate.}
    \label{fig:cross_metrics_etth1}
\end{figure}

\autoref{fig:cross_metrics_etth1} presents the comparison of ETTh1 performance before and after post-training across multiple metrics.
Notably, MSE-based metrics (RMSE, NRMSE) show consistent improvements, while absolute error metrics such as MAE, MAPE, and SMAPE exhibit slight deterioration.
This highlights the importance of aligning the optimization metric with the target deployment objectives, as post-training correction is most effective when guided by the metric that best reflects the desired outcomes.

\subsection{Human-in-the-loop illustration}

Finally, we illustrate the human-in-the-loop (HITL) functionality of our framework, where a user provides natural-language feedback that is translated into an executable correction function by a large language model (LLM).
\begin{figure}[t]
    \centering
    \includegraphics[width=0.6\linewidth]{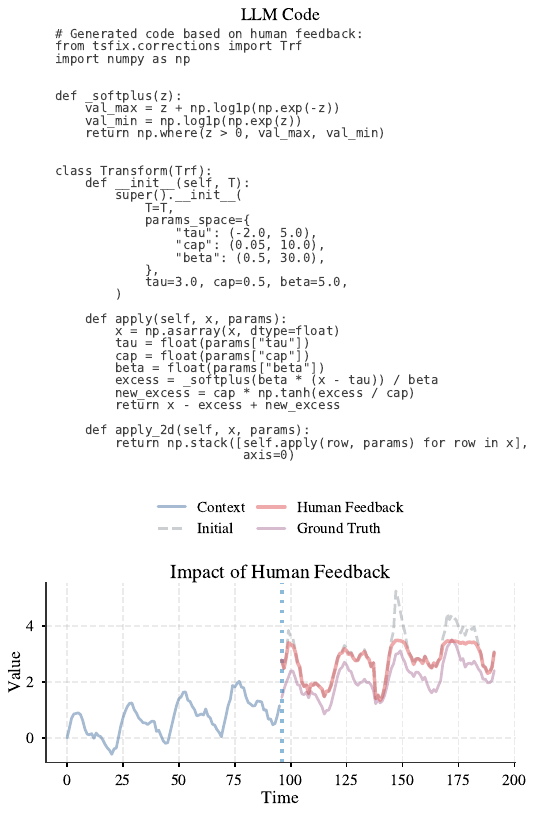}
    \caption{Illustration of human feedback post-training correction (left) along with the corresponding LLM generated code (right).}
    \label{fig:illus_human_feedback}
\end{figure}

In \autoref{fig:illus_human_feedback}, the left panel shows how the user-provided feedback modifies the forecast trajectory relative to the initial prediction.
The right panel shows the corresponding code generated by the LLM to implement the transformation. As in the rest of our framework, human-proposed corrections are treated as candidate actions and are accepted only if they improve the validation metric. This feature demonstrates the practical applicability of integrating expert feedback into the correction process, making the approach more flexible and customizable.

\section{Discussion}
\label{sec:discussion}

We discuss key practical considerations and trade-offs of our post-training strategy:

\begin{enumerate}[itemsep=0.1pt, topsep=0.1pt]
    \item \emph{Small-Model Improvements}: \quad Our approach is effective for smaller models, but the correction process can become a significant overhead if the base model's training time is short, making the correction time comparable to or greater than the training time.
    
    \item \emph{Beyond Time-Series}: \quad Although designed for time-series forecasting, our framework is generalizable to other tasks like regression, classification, and reinforcement learning, making it broadly applicable.

    \item \emph{Expressivity Ceiling}: \quad The benefits of post-training correction are more noticeable for simpler or less-optimized forecasters. For highly expressive models, like Crossformer, the gains from correction are more limited.

    \item \emph{Correction Set vs. Optimization Time}: \quad A larger correction set can improve performance, but this increases optimization time. If base training is quick, the post-training process can dominate the overall runtime, requiring careful balancing between gains and time costs.
\end{enumerate}
\section{Conclusion}
\label{sec:conclusion}

We introduced a model-agnostic post-training framework that improves time-series forecasts without retraining the base model. By selecting corrective actions within a fixed validation budget and optionally incorporating human feedback through a natural-language interface, the method consistently enhances accuracy with minimal runtime. 

Our experiments show significant improvements, with Successive Halving as a reliable budget-aware strategy. The theoretical analysis explains how the selected correction converges to the optimal action as the evaluation budget increases.

Beyond performance gains, the framework is practical for deployment: corrections are interpretable, auditable, and can be updated without retraining, making it ideal for scenarios where retraining is expensive or risky.

Future work will focus on richer families of corrections, improved LLM-to-action alignment, and extending the framework to streaming and multimodal settings with adaptive correction. A promising avenue is adapting the approach to image generative models, enabling simple user- or model-driven corrections that improve perceptual quality and mitigate common artifacts (\eg distortions, and aberrations).

\clearpage
\newpage

  
\bibliography{bibliography}
\bibliographystyle{tmlr}

\clearpage
\newpage
\appendix

\restoreTheoremLikeNumbers

\section{Appendix: table of contents}
\label{sec:supp_table_of_contents}

\begin{enumerate}[align=left,itemsep=0.5cm,label=\Alph*]
    \item Appendix: Notations summary \dotfill \pageref{sec:supp_notations}
    \item Appendix: Experimental settings \dotfill \pageref{sec:supp_exp_details}
    \item Appendix: Theoretical analysis \dotfill \pageref{sec:supp_theory}
\end{enumerate}

\section{Appendix: Notations summary}
\label{sec:supp_notations}

\renewcommand{\arraystretch}{1.5}

\begin{longtable}{@{}p{0.23\textwidth} p{0.73\textwidth}@{}}
    \textbf{Symbol}                                                       & \textbf{Description}                                                                       \\
    \hline
    \endfirsthead
    \textbf{Symbol}                                                       & \textbf{Description}                                                                       \\
    \hline
    \endhead
    \((X,Y)\sim\mathcal{D}\)                                              & Input-target random pair sampled from the forecasting data distribution.                   \\
    \hline
    \(f,\ f_\theta\)                                                      & Base forecaster (parameterized by \(\theta\)); frozen after training.                      \\
    \hline
    \(f_\lambda\)                                                         & Ridge forecaster with regularization parameter \(\lambda\).                                \\
    \hline
    \(Y_{\mathrm{true}}\)                                                 & Ground-truth target random variable.                                                       \\
    \hline
    \(Z:=f_\theta(X)\)                                                    & Scalar prediction of the base forecaster for input \(X\).                                  \\
    \hline
    \(g_{a,b}(z)=a\,z+b\)                                                 & Affine correction function with slope \(a\) and intercept \(b\).                           \\
    \hline
    \(Y_{\mathrm{corr}}=a\,f(X)+b\)                                       & Corrected prediction after applying the affine corrector to the base forecast.             \\
    \hline
    \(a^\star,\ b^\star\)                                                 & Optimal affine parameters minimizing squared risk in the affine family.                    \\
    \hline
    \(R(a,b):=\mathbb{E}\!\left[(g_{a,b}(Z)-Y_{\mathrm{true}})^2\right]\) & Population squared risk of the affine-corrected predictor.                                 \\
    \hline
    \(R_0:=\mathbb{E}\!\left[(Z-Y_{\mathrm{true}})^2\right]\)             & Uncorrected base risk.                                                                     \\
    \hline
    \(R^\star:=\min_{a,b}R(a,b)\)                                         & Optimal corrected risk in the affine family.                                               \\
    \hline
    \(\Delta R:=R_0-R^\star\)                                             & Affine correction gain (non-negative).                                                     \\
    \hline
    \(\{g_k\}_{k=1}^{K}\)                                                 & Discrete set of candidate correction functions (arms/actions).                             \\
    \hline
    \(K\)                                                                 & Number of candidate correction functions.                                                  \\
    \hline
    \(g_{\mathrm{h}}\)                                                    & Optional human-defined correction generated from expert feedback (via LLM mediation).      \\
    \hline
    \(g_{k,\beta}\)                                                       & Correction \(k\) with continuous hyperparameters \(\beta\).                                \\
    \hline
    \(\beta_k^\star\)                                                     & Oracle hyperparameter for arm \(k\), minimizing true risk.                                 \\
    \hline
    \(M\)                                                                 & Number of correction steps composed in the post-training pipeline.                         \\
    \hline
    \(h=g_{k_M}\circ\cdots\circ g_{k_1}\circ f\)                          & Final corrected forecaster after composing \(M\) selected corrections with the base model. \\
    \hline
    \(h^\star\)                                                           & Best corrected forecaster returned by the search/selection procedure.                      \\
    \hline
    \(T\)                                                                 & Total evaluation budget for the bandit selection procedure.                                \\
    \hline
    \(L_{k,t}\in[0,L_{\max}]\)                                            & Clipped squared loss observed when evaluating arm \(k\) at pull \(t\).                     \\
    \hline
    \(L_{\max}\)                                                          & Clipping constant used in the bounded-loss assumption.                                     \\
    \hline
    \(R_k:=R(g_{k,\beta_k^\star}\circ f_\theta)=\mathbb{E}[L_{k,t}]\)     & True (clipped) risk of arm \(k\) under its oracle hyperparameter.                          \\
    \hline
    \(k^\star\in\arg\min_{k\in[K]}R_k\)                                   & Optimal corrective action index.                                                           \\
    \hline
    \(k_T\)                                                               & Arm selected by Successive Halving after exhausting budget \(T\).                          \\
    \hline
    \(\Delta_k:=R_k-R_{k^\star}\)                                         & Suboptimality gap of arm \(k\).                                                            \\
    \hline
    \(\Delta_{\max}:=\max_k\Delta_k\)                                     & Largest gap among candidate corrections.                                                   \\
    \hline
    \(R:=\lceil\log_2 K\rceil\)                                           & Number of Successive Halving rounds.                                                       \\
    \hline
    \(n_r:=\left\lfloor \frac{T}{K R}2^r\right\rfloor\)                   & Evaluations per surviving arm at SH round \(r\).                                           \\
    \hline
    \(n_0=\left\lfloor\frac{T}{K R}\right\rfloor\)                        & First-round per-arm evaluation count.                                                      \\
    \hline
    \(\Delta:=R_{k^-}-R_{k^\star}\) (for \(K=2\))                         & Two-arm risk gap used in corollary bounds.                                                 \\
    \hline
    \(\widehat{R}_{k,r}\)                                                 & Empirical risk estimate of arm \(k\) at SH round \(r\).                                    \\
    \hline
\end{longtable}

\section{Appendix: Experimental settings}
\label{sec:supp_exp_details}

In this section, we provide additional details on the experimental setting.

\subsection{Correction set}

In our experiments, we consider the following default set of correction functions (applied to the forecaster's outputs):
\begin{itemize}
    \item \texttt{scale\_amplitude}: globally rescales the forecast around its mean to increase/decrease oscillation amplitude.
    \item \texttt{piecewise\_scale\_high}: rescales only the upper part of the forecast to adjust peaks.
    \item \texttt{piecewise\_scale\_low}: rescales only the lower part of the forecast to adjust troughs.
    \item \texttt{add\_linear\_trend\_slope}: adds a linear trend with tunable slope to correct systematic drift over the horizon.
    \item \texttt{add\_linear\_trend\_intercept}: adds a constant offset to correct global bias.
    \item \texttt{increase\_minimum\_factor}: increases the minimum level of the forecast to avoid underestimation of lows.
    \item \texttt{increase\_maximum\_factor}: increases the maximum level of the forecast to avoid underestimation of highs.
\end{itemize}

\subsection{Datasets}

The following table outlines the dataset names, their sources, key characteristics, and the corresponding references for the papers that describe each dataset.

\begin{longtable}[c]{@{}>{\raggedright\arraybackslash}p{3.5cm}>{\raggedright\arraybackslash}p{4cm}>{\raggedright\arraybackslash}p{5.5cm}@{}}
    \caption{Datasets Overview}                                                                                                                                                                                             \\
    \toprule
    \rowcolor{gray!20} \textbf{Dataset Name} & \textbf{Source and Reference}                        & \textbf{Characteristics}                                                                                              \\
    \midrule
    \endfirsthead
    \multicolumn{3}{c}{\textbf{Datasets Overview (Continued)}}                                                                                                                                                              \\
    \toprule
    \rowcolor{gray!20} \textbf{Dataset Name} & \textbf{Source and Reference}                        & \textbf{Characteristics}                                                                                              \\
    \midrule
    \endhead
    \bottomrule
    \endfoot
    \texttt{ETTh1}                           & ETTh (Electricity) Benchmark \cite{zhou2021informer} & 1-hour-level time-series with 6 features and "oil temperature" as the target. Train/val/test split: 12/4/4 months.    \\[2ex]
    \texttt{ETTh2}                           & ETTh (Electricity) Benchmark \cite{zhou2021informer} & 1-hour-level time-series with 6 features and "oil temperature" as the target. Includes more features than ETTh1.      \\[2ex]
    \texttt{ETTm1}                           & ETTh (Electricity) Benchmark \cite{zhou2021informer} & 15-minute-level time-series with 6 features and "oil temperature" as the target. Train/val/test split: 12/4/4 months. \\[2ex]
    \texttt{ETTm2}                           & ETTh (Electricity) Benchmark \cite{zhou2021informer} & 15-minute-level time-series, similar to ETTm1, with different subsets for long-term forecasting.                      \\[2ex]
    \texttt{KDD Cup}                         & Open TS Benchmark \cite{qiu2024tfb}                  & Long hourly time series representing the air quality levels in 59 stations in Beijing and London.                     \\[2ex]
    \texttt{Pedestrian}                      & Open TS Benchmark \cite{qiu2024tfb}                  & Pedestrian count data from urban settings, used for mobility prediction.                                              \\[2ex]
    \texttt{Exchange Rate}                   & Open TS Benchmark \cite{qiu2024tfb}                  & Daily exchange-rate series for eight countries/currencies.                                                            \\[2ex]
\end{longtable}

\section{Appendix: Theoretical analysis}
\label{sec:supp_theory}

This appendix provides detailed proofs for the main theoretical results stated in \autoref{sec:theory}.

\subsection{Proof of~\autoref{THM:AFFINE_CORRECTION}}

\begin{theoremwithtag}{A}[Optimal affine correction of a forecaster]
    For $(X, Y_{\mathrm{true}}) \sim \mathcal{D}$, with $a^\star$ and $b^\star$ defined above, the affine-corrected risk is:
    \[
        R(a^\star,b^\star)=\Var(Y_{\mathrm{true}})-\frac{\Cov(Y_{\mathrm{true}},Z)^2}{\Var(Z)}.
    \]
    and the risk reduction satisfies
    \[
        R_0-R^\star = \big(\E[Y_{\mathrm{true}}]-\E[Z]\big)^2 + \frac{\big(\Var(Z)-\Cov(Y_{\mathrm{true}},Z)\big)^2}{\Var(Z)}.
    \]
\end{theoremwithtag}

\begin{proof}

    Let $Z:=f_\theta(X)$ and write $m_Z:=\E[Z]$, $m_Y:=\E[Y_{\mathrm{true}}]$. Define
    \[
        R(a,b):=\E\!\left[(aZ+b-Y_{\mathrm{true}})^2\right].
    \]

    Fix $a\in\R$. Viewing $R(a,b)$ as a quadratic in $b$, differentiate:
    \[
        \frac{\partial}{\partial b}R(a,b)=2\,\E[aZ+b-Y_{\mathrm{true}}].
    \]
    Setting this to $0$ gives the optimal intercept
    \[
        b^\star(a)=m_Y-a m_Z.
    \]
    Plugging back,
    \[
        R(a,b^\star(a))=\E\!\left[\big(a(Z-m_Z)-(Y_{\mathrm{true}}-m_Y)\big)^2\right].
    \]
    Expanding and using $\E[Z-m_Z]=\E[Y_{\mathrm{true}}-m_Y]=0$,
    \[
        R(a,b^\star(a))=a^2\Var(Z)-2a\,\Cov(Z,Y_{\mathrm{true}})+\Var(Y_{\mathrm{true}}).
    \]
    If $\Var(Z)>0$, this quadratic is minimized at
    \[
        a^\star=\frac{\Cov(Z,Y_{\mathrm{true}})}{\Var(Z)},\qquad
        b^\star=m_Y-a^\star m_Z,
    \]
    and the optimal value is
    \[
        R^\star
        = \Var(Y_{\mathrm{true}})-\frac{\Cov(Z,Y_{\mathrm{true}})^2}{\Var(Z)}.
    \]

    For the risk reduction, let $R_0:=\E[(Z-Y_{\mathrm{true}})^2]$. Decompose $Z-Y_{\mathrm{true}}=(m_Z-m_Y)+\big((Z-m_Z)-(Y_{\mathrm{true}}-m_Y)\big)$ to get
    \[
        R_0=(m_Z-m_Y)^2+\Var(Z)+\Var(Y_{\mathrm{true}})-2\Cov(Z,Y_{\mathrm{true}}).
    \]
    Hence
    \[
        R_0-R^\star
        =(m_Y-m_Z)^2+\Var(Z)-2\Cov(Z,Y_{\mathrm{true}})
        +\frac{\Cov(Z,Y_{\mathrm{true}})^2}{\Var(Z)}
        =(m_Y-m_Z)^2+\frac{(\Var(Z)-\Cov(Z,Y_{\mathrm{true}}))^2}{\Var(Z)}\ge 0.
    \]

\end{proof}

\subsection{Proof of~\autoref{THM:UPPER_BOUND_RISK}}

\begin{theoremwithtag}{B}[Corrected prediction risk upper bound]

    Assume $M=1$ and let $K\ge 2$ be the number of corrective actions.
    Let $R_k := R(g_{k,\beta_k^\star}\circ f_\theta)$ and let
    \mbox{\small\(k^\star \in \arg\min_{k\in[K]} R_k\)},
    \mbox{\small\(\Delta_k := R_k - R_{k^\star}\ \ (\ge 0)\)},
    \mbox{\small\(\Delta_{\max} := \max_{k\in[K]}\Delta_k\)}.
    Run \emph{Successive Halving} (SH) with total evaluation budget $T$ and $R:=\lceil \log_2 K\rceil$ rounds as follows:
    at round $r\in\{0,\dots,R-1\}$, each surviving arm is evaluated $n_r$ times, where
        {\small
            \[
                n_r := \left\lfloor \frac{T}{K R}\,2^r \right\rfloor,
            \]
        }
    and then the worst half of the arms (largest empirical risks) are eliminated.
    Let $k_T$ be the arm returned by SH at the end. Under~\autoref{ass:bounded_sqerr}, the expected corrected prediction risk satisfies
        {\small
            \[
                \mathbb{E}\!\left[ R\!\big(g_{k_T,\beta_{k_T}^\star}\circ f_\theta\big) \right]
                \le
                R_{k^\star}
                +
                2\,\Delta_{\max}\sum_{r=0}^{R-1}\ \sum_{k\neq k^\star}
                \exp\!\left(-\frac{n_r\,\Delta_k^2}{2L_{\max}^2}\right).
            \]
        }
    In particular, since $n_r\ge n_0$ for all $r$ and $R=\lceil\log_2 K\rceil$,
    {\small
            \[
                \mathbb{E}\!\left[ R\!\big(g_{k_T,\beta_{k_T}^\star}\circ f_\theta\big) \right]
                \le
                R_{k^\star}
                +
                2\,\Delta_{\max}\,R\sum_{k\neq k^\star}
                \exp\!\left(-\frac{n_0\,\Delta_k^2}{2L_{\max}^2}\right),
            \]
        }
    where
        {\small
            \[
                n_0=\left\lfloor \frac{T}{K R}\right\rfloor.
            \]
        }
\end{theoremwithtag}

\begin{proof}
    We prove the bound by decomposing the expectation and then controlling the probability that SH returns a suboptimal arm.

    Fix a round $r$. Let $S_r\subseteq[K]$ be the set of surviving arms at the beginning of round $r$ (so $k^\star\in S_r$ until it is eliminated).
    For each surviving arm $k\in S_r$, SH collects $n_r$ i.i.d.\ losses $L_{k,1},\dots,L_{k,n_r}$ and forms the empirical risk
    \[
        \widehat{R}_{k,r}
        := \frac{1}{n_r}\sum_{i=1}^{n_r} L_{k,i}.
    \]
    By~\autoref{ass:bounded_sqerr}, each $L_{k,i}\in[0,L_{\max}]$ and $\mathbb{E}[L_{k,i}]=R_k$.

    Write the returned arm as $k_T$. Since $k^\star$ minimizes the true risk, we have for every outcome:
    \[
        R_{k_T}
        =
        R_{k^\star} + (R_{k_T}-R_{k^\star})
        =
        R_{k^\star} + \Delta_{k_T}.
    \]
    If SH returns the optimal arm ($k_T=k^\star$), then $\Delta_{k_T}=0$. If SH returns a suboptimal arm ($k_T\neq k^\star$), then $\Delta_{k_T}\le \Delta_{\max}$ by definition of $\Delta_{\max}$.
    Therefore,
    \[
        \Delta_{k_T}
        \le
        \Delta_{\max}\,\mathbf{1}\{k_T\neq k^\star\}.
    \]
    Taking expectations and substituting back gives
    \begin{equation}
        \label{eq:step1_decomp}
        \mathbb{E}[R_{k_T}]
        =
        R_{k^\star} + \mathbb{E}[\Delta_{k_T}]
        \le
        R_{k^\star} + \Delta_{\max}\,\mathbb{P}(k_T\neq k^\star).
    \end{equation}
    Thus, it remains to upper bound the probability of returning a suboptimal arm: $\mathbb{P}(k_T\neq k^\star)$.

    Fix a round $r$ and suppose $k^\star\in S_r$ at the start of this round. SH eliminates the \emph{worst half} of the arms according to $\widehat{R}_{k,r}$.
    Hence, for $k^\star$ to be eliminated at round $r$, at least half of the other surviving arms must have empirical risk no larger than that of $k^\star$.
    In particular, a necessary condition is that \emph{there exists at least one competitor} $k\in S_r\setminus\{k^\star\}$
    such that
    \[
        \widehat{R}_{k,r} \le \widehat{R}_{k^\star,r}.
    \]
    Therefore the elimination event satisfies the set inclusion
    \begin{equation}
        \label{eq:step2_inclusion}
        \{k^\star \text{ is eliminated at round } r\}
        \subseteq
        \bigcup_{k\in S_r\setminus\{k^\star\}} \{\widehat{R}_{k,r} \le \widehat{R}_{k^\star,r}\}.
    \end{equation}

    If SH outputs a suboptimal arm ($k_T\neq k^\star$), then $k^\star$ must have been eliminated in \emph{some} round. Thus,
    \[
        \{k_T\neq k^\star\}
        \subseteq
        \bigcup_{r=0}^{R-1} \{k^\star \text{ is eliminated at round } r\}.
    \]
    Apply the union bound, then use \eqref{eq:step2_inclusion}, and apply the union bound again:
    \begin{align}
        \mathbb{P}(k_T\neq k^\star)
         & \le
        \sum_{r=0}^{R-1}\mathbb{P}(k^\star \text{ is eliminated at round } r)
        \nonumber \\
         & \le
        \sum_{r=0}^{R-1}\ \sum_{k\in S_r\setminus\{k^\star\}}
        \mathbb{P}\big(\widehat{R}_{k,r} \le \widehat{R}_{k^\star,r}\big)
        \nonumber \\
         & \le
        \sum_{r=0}^{R-1}\ \sum_{k\neq k^\star}
        \mathbb{P}\big(\widehat{R}_{k,r} \le \widehat{R}_{k^\star,r}\big).
        \label{eq:step3_union}
    \end{align}
    In the last line, we used $S_r\setminus\{k^\star\}\subseteq [K]\setminus\{k^\star\}$ to simplify the sum.

    So it remains to bound, for each fixed pair $(k,r)$ with $k\neq k^\star$,
    \[
        \mathbb{P}\big(\widehat{R}_{k,r} \le \widehat{R}_{k^\star,r}\big).
    \]

    Fix $k\neq k^\star$ and round $r$. Recall $\Delta_k = R_k-R_{k^\star}>0$. If $\widehat{R}_{k,r}\le \widehat{R}_{k^\star,r}$, then
    \[
        (\widehat{R}_{k,r}-R_k) - (\widehat{R}_{k^\star,r}-R_{k^\star})
        \le
        -(R_k-R_{k^\star})
        =
        -\Delta_k.
    \]
    Using the implication:
    \[
        \widehat{R}_{k,r}\le \widehat{R}_{k^\star,r}
        \ \Longrightarrow\
        \Big(\widehat{R}_{k,r}\le R_k-\tfrac{\Delta_k}{2}\Big)
        \ \ \text{or}\ \
        \Big(\widehat{R}_{k^\star,r}\ge R_{k^\star}+\tfrac{\Delta_k}{2}\Big).
    \]
    Indeed, if \emph{neither} of these two events holds, then simultaneously $\widehat{R}_{k,r} > R_k-\Delta_k/2$ and $\widehat{R}_{k^\star,r} < R_{k^\star}+\Delta_k/2$, and subtracting yields
    \[
        \widehat{R}_{k,r}-\widehat{R}_{k^\star,r}
        >
        (R_k-\tfrac{\Delta_k}{2})-(R_{k^\star}+\tfrac{\Delta_k}{2})
        =
        R_k-R_{k^\star}-\Delta_k
        =
        0,
    \]
    which contradicts $\widehat{R}_{k,r}\le \widehat{R}_{k^\star,r}$.

    Therefore, by a union bound,
    \begin{equation}
        \label{eq:step4_split}
        \mathbb{P}\big(\widehat{R}_{k,r} \le \widehat{R}_{k^\star,r}\big)
        \le
        \mathbb{P}\!\left(\widehat{R}_{k,r}\le R_k-\tfrac{\Delta_k}{2}\right)
        +
        \mathbb{P}\!\left(\widehat{R}_{k^\star,r}\ge R_{k^\star}+\tfrac{\Delta_k}{2}\right).
    \end{equation}

    Because each $L_{k,i}\in[0,L_{\max}]$ and $\widehat{R}_{k,r}$ is an average of $n_r$ i.i.d.\ samples with mean $R_k$, Hoeffding's inequality gives, for any $\varepsilon>0$,
    \[
        \mathbb{P}\big(\widehat{R}_{k,r}-R_k \le -\varepsilon\big)
        \le
        \exp\!\left(-\frac{2n_r\varepsilon^2}{L_{\max}^2}\right),
        \qquad
        \mathbb{P}\big(\widehat{R}_{k,r}-R_k \ge \varepsilon\big)
        \le
        \exp\!\left(-\frac{2n_r\varepsilon^2}{L_{\max}^2}\right).
    \]
    Apply these bounds to \eqref{eq:step4_split} with $\varepsilon=\Delta_k/2$. This yields, for every $k\neq k^\star$ and every round $r$,
    \begin{align}
        \mathbb{P}\big(\widehat{R}_{k,r} \le \widehat{R}_{k^\star,r}\big)
         & \le
        \exp\!\left(-\frac{2n_r(\Delta_k/2)^2}{L_{\max}^2}\right)
        +
        \exp\!\left(-\frac{2n_r(\Delta_k/2)^2}{L_{\max}^2}\right)
        \nonumber \\
         & =
        2\exp\!\left(-\frac{n_r\,\Delta_k^2}{2L_{\max}^2}\right).
        \label{eq:step5_pairwise}
    \end{align}

    Use \eqref{eq:step5_pairwise} into \eqref{eq:step3_union}:
    \[
        \mathbb{P}(k_T\neq k^\star)
        \le
        \sum_{r=0}^{R-1}\ \sum_{k\neq k^\star}
        2\exp\!\left(-\frac{n_r\,\Delta_k^2}{2L_{\max}^2}\right)
        =
        2\sum_{r=0}^{R-1}\ \sum_{k\neq k^\star}
        \exp\!\left(-\frac{n_r\,\Delta_k^2}{2L_{\max}^2}\right).
    \]
    Finally, substitute this into \eqref{eq:step1_decomp}:
    \[
        \mathbb{E}[R_{k_T}]
        \le
        R_{k^\star}
        +
        \Delta_{\max}\cdot
        2\sum_{r=0}^{R-1}\ \sum_{k\neq k^\star}
        \exp\!\left(-\frac{n_r\,\Delta_k^2}{2L_{\max}^2}\right),
    \]
    which is exactly the first displayed inequality in~\autoref{THM:UPPER_BOUND_RISK}. Moreover, since $n_r$ is nondecreasing in $r$, we have $n_r\ge n_0$ for all $r$.
    Because $x\mapsto e^{-cx}$ is nonincreasing for $c>0$, we obtain
    \[
        \exp\!\left(-\frac{n_r\,\Delta_k^2}{2L_{\max}^2}\right)
        \le
        \exp\!\left(-\frac{n_0\,\Delta_k^2}{2L_{\max}^2}\right).
    \]
    Summing over $r=0,\dots,R-1$ yields
    \[
        \sum_{r=0}^{R-1}\exp\!\left(-\frac{n_r\,\Delta_k^2}{2L_{\max}^2}\right)
        \le
        R\exp\!\left(-\frac{n_0\,\Delta_k^2}{2L_{\max}^2}\right),
    \]
    and substituting this back gives the second displayed inequality of~\autoref{THM:UPPER_BOUND_RISK}.

\end{proof}

\end{document}